\documentclass[11pt,a4paper]{article}
\usepackage[hyperref]{acl2021}
\usepackage{times}
\usepackage{amsmath}
\usepackage{amssymb}
\usepackage{amsthm}
\usepackage{autobreak}
\usepackage{mathrsfs}
\usepackage{latexsym}
\usepackage{booktabs}
\usepackage{graphicx}
\usepackage{subfigure}
\usepackage{footnote}
\usepackage{color}
\usepackage{epstopdf}

\usepackage{CJK}
\usepackage{url}
\usepackage{float}
\usepackage{microtype}
\usepackage{makecell}
\usepackage{multirow}

\aclfinalcopy 

\title{Knowledge-Empowered Representation Learning for Chinese Medical Reading Comprehension: Task, Model and Resources}

\author{Taolin Zhang$^{1,2}$\thanks{\ \ T. Zhang and C. Wang contributed equally to this work.}, Chengyu Wang$^{2}$\footnotemark[1], Minghui Qiu$^{2}$, Bite Yang$^{3}$ \\ \textbf{Zerui Cai}$^{4}$, \textbf{Xiaofeng He}$^{4}$\thanks{\ \ Corresponding author.}, \textbf{Jun Huang}$^{2}$\\
$^1$ School of Software Engineering, East China Normal University
$^2$ Alibaba Group\\
$^3$ DXY
$^4$ School of Computer Science and Technology, East China Normal University\\
 {\tt  zhangtl0519@gmail.com, zrcai\_flow@126.com, yangbt@dxy.cn}\\
 {\tt \{chengyu.wcy, minghui.qmh, huangjun.hj\}@alibaba-inc.com} \\
 {\tt  hexf@cs.ecnu.edu.cn}}

\date{}

\begin{document}
\maketitle
\begin{CJK}{UTF8}{gbsn}
\begin{abstract}
Machine Reading Comprehension (MRC) aims to extract answers to questions given a passage, which has been widely studied recently especially in open domains. However, few efforts have been made on closed-domain MRC, mainly due to the lack of large-scale training data. In this paper, we introduce a multi-target MRC task for the medical domain, whose goal is to predict answers to medical questions and the corresponding support sentences from medical information sources simultaneously, in order to ensure the high reliability of medical knowledge serving. A high-quality dataset (more than 18k samples) is manually constructed for the purpose, named Multi-task \textbf{C}hinese \textbf{Med}ical \textbf{MRC} dataset (CMedMRC), with detailed analysis conducted. We further propose a \textbf{C}hinese \textbf{med}ical \textbf{BERT} model for the task (CMedBERT), which fuses medical knowledge into pre-trained language models by the dynamic fusion mechanism of heterogeneous features and the multi-task learning strategy.
Experiments show that CMedBERT consistently outperforms strong baselines by fusing context-aware and knowledge-aware token representations.\footnote{The code and dataset will be available at \url{https://github.com/MatNLP/CMedMRC}}
\end{abstract}


\section{Introduction}
Machine Reading Comprehension (MRC) has become a popular task in NLP, aiming to understand a given passage and answer the relevant questions. With the wide availability of MRC datasets~\citep{DBLP:conf/emnlp/RajpurkarZLL16, DBLP:conf/acl/HeLLLZXLWWSLWW18, DBLP:conf/emnlp/CuiLCXCMWH19} and deep learning models~\citep{DBLP:conf/iclr/YuDLZ00L18, DBLP:conf/acl/DingZCYT19} (including pre-trained language models such as BERT \citep{DBLP:conf/naacl/DevlinCLT19}), significant progress has been made.

\begin{figure}
\centering
\includegraphics[height=8cm,width=7cm]{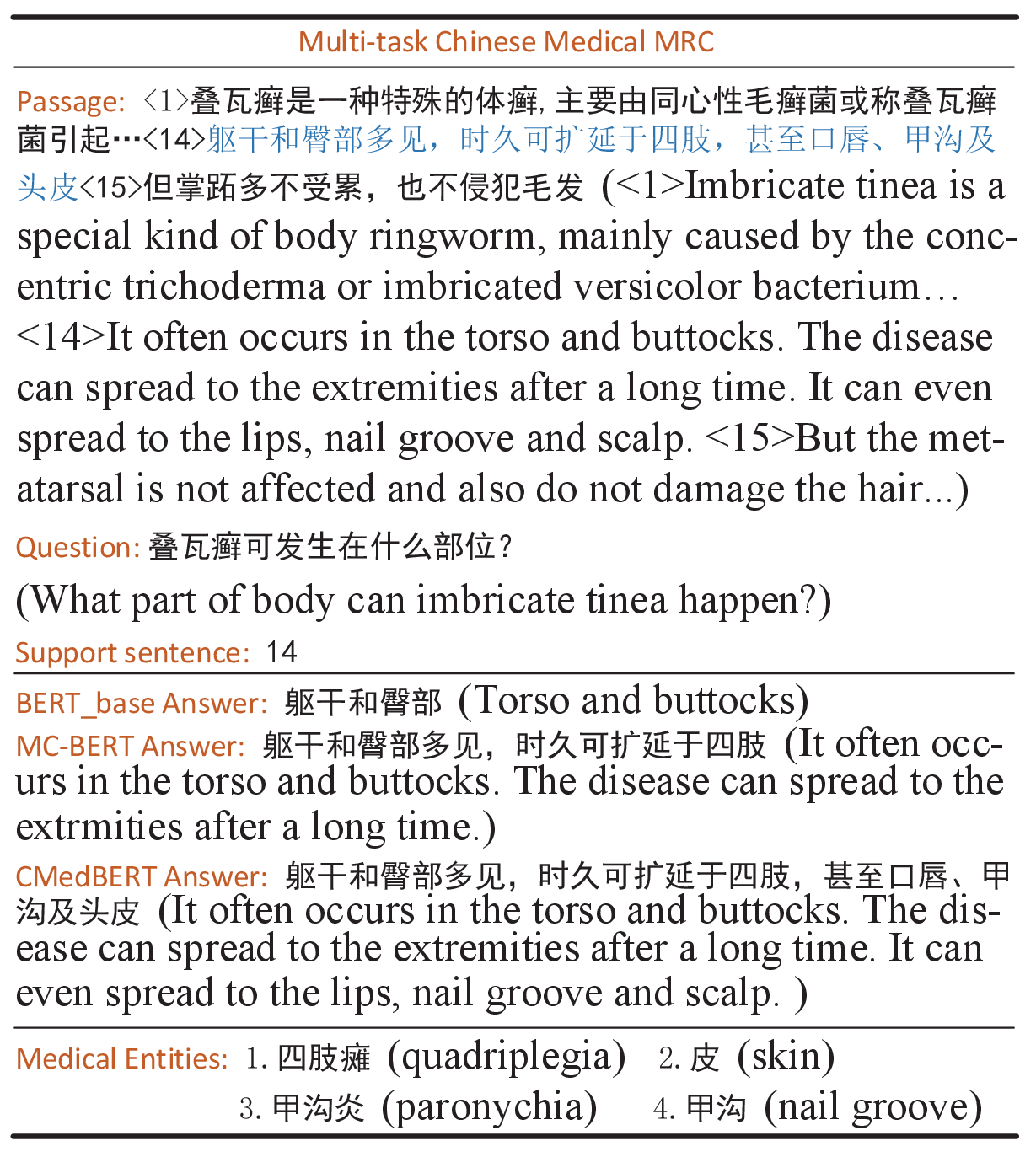}
\caption{Dataset example. MC-BERT and BERT can only predict part of the correct answer. With medical knowledge fused, our CMedBERT model can extract the complete answer. Contents in brackets refer to English translations.}
\label{figure_motivation_example}
\end{figure}

Despite the success, a majority of MRC research has focused on open domains.
For specific domains, however, the construction of high-quality MRC datasets, together with the design of corresponding models is considerably deficient~\citep{DBLP:conf/aclnut/WelblLG17, DBLP:journals/tacl/WelblSR18}. The causes behind this phenomenon are threefold. Take the medical domain as an example.
i) Data annotators are required to have medical backgrounds with high standards. Hence, simple crowd-sourcing~\citep{DBLP:conf/emnlp/RajpurkarZLL16, DBLP:conf/emnlp/CuiLCXCMWH19} often leads to poor annotation results. ii) Due to the domain sensitivity, people are more concerned about the reliability of the information sources where the answers are extracted, and the explainability of the answers themselves~\citep{lee2014interventions, dalmer2017questioning}. This is fundamentally different from the task requirements of open-domain MRC. iii) From the perspective of model learning, it is difficult for pre-trained language models to understand the meaning of the questions and passages containing a lot of specialized terms~\citep{DBLP:conf/acl/ChenBM16, DBLP:conf/emnlp/BauerWB18}. Without the help of domain knowledge, state-of-the-art models can perform poorly.
As shown in Figure~\ref{figure_motivation_example}, BERT~\citep{DBLP:conf/naacl/DevlinCLT19} and MC-BERT ~\citep{DBLP:journals/corr/abs-2008-10813} only predict part of the correct answer, i.e.,~``torso" and ``buttocks", instead of generating the complete answer to the medical question.

In this paper, we present a comprehensive study on Chinese medical MRC, including i) how the task is formulated, ii) the construction of the Chinese medical dataset and iii) the MRC model with rich medical knowledge injected. To meet the requirements of medical MRC, we aim to predict both the answer spans to a medical question, and the support sentence from the passage, indicating the source of the answer. The support sentences provide abundant evidence for users to learn medical knowledge, and for medical professionals to assess the trustworthiness of model output results.

For the dataset, we construct a highly-quality Chinese medical MRC dataset, named the Multi-task Chinese Medical MRC dataset (CMedMRC). It contains 18,153 $<$\textit{question, passage, answer, support sentence}$>$ quads. Based on the analysis of CMedMRC, we summarize four special challenges for Chinese medical MRC, including \textit{long-tail terminologies, synonym terminology, terminology combination} and \textit{paraphrasing}.
In addition, we find that comprehensive skills are required for MRC models to answer medical questions correctly.
For answer extraction in CMedMRC, direct \textit{token matching} is required for answering 31\% of the questions, \textit{co-reference resolution} for 11\%, \textit{multi-sentence reasoning} for 18\% and \textit{implicit causality} for 22\%. In addition, the answers to the remaining questions (16\%) are extremely difficult to extract without rich medical background knowledge.

To address the medical MRC task, we propose the multi-task dynamic heterogeneous fusion network (CMedBERT) based on the MC-BERT~\citep{DBLP:journals/corr/abs-2008-10813} model and a Chinese medical knowledge base (see Appendix). The technical contributions of CMedBERT are twofold:
\begin{itemize}
\item\textbf{Heterogeneous Feature Fusion}: We mimic humans' approach of reading comprehension~\citep{wang1999reading} by learning attentively  aggregated representations of multiple entities in the passage. Different from the knowledge fusion method used by KBLSTM ~\citep{DBLP:conf/acl/YangM17} and KT-NET ~\citep{DBLP:conf/acl/YangWLLLWSL19}, we propose a two-level attention and a gated-loop mechanism to replace the knowledge sentinel, so that rich knowledge representations can be better integrated into the model.
\item\textbf{Multi-task Learning}: Parameters of CMedBERT are dynamically learned by capturing the relationships between the two tasks via multi-task learning. We regard the semantic similarities between support sentences and answers to questions as the task similarities.
\end{itemize}


We compare CMedBERT against six strong baselines. For answer prediction, compared to the strongest competitor, the EM (Exact Match) and F1 scores are increased by +3.88\% and +1.46\%, respectively.
Meanwhile, the support sentence prediction task result is increased by a large margin, i.e., +7.81\% of EM and +4.07\% of F1.





\begin{table*}[]
\scriptsize
\centering
\begin{tabular}{clc}
\toprule
\textbf{Challenges} & \textbf{Characteristics}  & \textbf{Example} \\ \cline{1-3}
 & \textbf{\textit{Long-tail terminology}}  & \makecell*[l]{冈上肌肌腱断裂试验是对冈上肌肌腱是否存在断裂进行检查。 (\underline{{\color{blue} \textit{supraspinatus tendon}}} rupture \\test is to check whether the supraspinatus tendon is ruptured.)}    \\ \cline{2-3}
\textbf{Lexical-Level}    & \textbf{\textit{Synonym terminology}}  & \makecell*[l]{...本药品对过敏性鼻炎和上呼吸道感染引起的鼻充血有效，可用于感冒或鼻窦炎... \\ (...This medicine is effective for nasal congestion caused by allergic rhinitis and \underline{\color{blue} \textit{upper respiratory tract}} \\ \underline{\color{blue} \textit{infection}}, and can be used for \underline{\color{blue} \textit{colds}} or sinusitis...)}  \\ \cline{2-3} 
    & \textbf{\textit{Terminology combination}}  & \makecell*[l]{...糖尿病性视网膜病(diabetic retinopathy)是糖尿病性微血管病变中最重要的表现... \\ (...\underline{\color{blue} \textit{Diabetic retinopathy (DR)}} is the most important manifestation of diabetic microangiopathy...)}
  \\ \cline{1-3}
\textbf{Sentence-Level}   & \textbf{\textit{Paraphrasing}}  & \makecell*[l]{\textbf{\textit{Passage}}:\tiny ...如果在嘴角烂了或结痂的地方进行冷敷，一方面冷敷物品不干净的话会造成感染；另一方面局部温\\度降低了之后，反而会延缓伤口的愈合。...(...If you apply a cold compress on a rotten or crusted \\corner of the mouth, on the one hand, if the cold compress is not clean, it will \underline{\color{blue} \textit{cause infection}}; on the \\other hand, when the local temperature is lowered, it will \underline{\color{blue} \textit{delay the healing of the wound}}...)\\\textbf{\textit{Question}}:\tiny 为什么嘴角烂了或结痂不建议进行冷敷？(Why is it \underline{\color{blue} \textit{not recommended}} to apply cold compresses when the \\corners of the mouth are rotten or crusted?)} \\ \bottomrule
\end{tabular}

\caption{\label{question_challenge_type} Two levels of challenges in processing Chinese medical texts. The blue and underscore contents in brackets indicate why this example belongs to its corresponding ``Characteristics'' category. (Best viewed in color.)}
\end{table*}

\section{Related Work}
\textbf{MRC Datasets and Models.} 
Due to the popularity of the MRC task, there exist many types of MRC datasets, such as span-extraction~\citep{DBLP:conf/emnlp/RajpurkarZLL16,DBLP:conf/emnlp/Yang0ZBCSM18}, multiple choices~\citep{DBLP:conf/emnlp/RichardsonBR13, DBLP:conf/emnlp/LaiXLYH17}, cloze-style~\citep{DBLP:conf/nips/HermannKGEKSB15}, cross-lingual ~\citep{DBLP:conf/emnlp/JingXZ19, DBLP:conf/acl/YuanSBGLDFJ20}.
For specific domains, however, the number of publicly available MRC datasets remains few, including SciQ~\citep{DBLP:conf/aclnut/WelblLG17}, Quasar-S~\citep{DBLP:journals/corr/DhingraMC17} and Biology~\citep{DBLP:conf/emnlp/BerantSCLHHCM14}.
CLiCR ~\citep{DBLP:conf/naacl/SusterD18} is a cloze-style single-task English medical MRC dataset. However, it contains a relatively small variety of medical questions, automatically generated from clinical case reports.
Recently, ~\citep{DBLP:conf/emnlp/LiHCPW20} propose a multi-choice Chinese medical QA dataset, retrieving text snippets as the passage and the task only chooses an existing correct option from candidate set.
Our work specifically focuses on the fine-grained medical MRC tasks and deep domain knowledge reasoning, with a manually constructed high-quality dataset released.

The model architecture of MRC mostly takes advantage of neural networks to learn token representations of passages and questions jointly~\citep{DBLP:journals/corr/abs-1906-03824, DBLP:journals/corr/abs-1907-01118}. The interaction between questions and passages is modeled based on attention mechanisms. 
The rapid development of deep learning leads to a variety of models, such as the QANet~\citep{DBLP:conf/iclr/YuDLZ00L18}, SAN~\citep{DBLP:conf/acl/GaoDLS18}.
Graph neural networks have been used in MRC recently by modeling the relations between entities in the passage~\citep{DBLP:conf/acl/DingZCYT19} and multi-grained tokens representation~\citep{DBLP:conf/acl/ZhengWLDCJZL20}.

\textbf{Pre-trained Language Models and Knowledge Fusion.} 
Pre-trained language models (e.g.,~BERT~\citep{DBLP:conf/naacl/DevlinCLT19}, ERNIE-THU \citep{DBLP:conf/acl/ZhangHLJSL19}, K-BERT~\citep{DBLP:conf/aaai/LiuZ0WJD020}) have successfully improved the performance of the MRC task, which even exceed the human level in some datasets.
This is because these models obtain better token representations and capture lexical and syntactic knowledge in different layers ~\citep{DBLP:conf/icml/GuanWZCH019}.
For specific domain, there also have some pre-trained models~\citep{DBLP:conf/emnlp/BeltagyLC19,DBLP:journals/corr/abs-2008-10813}.

A potential drawback is that pre-trained language models of open domains only learn general representations, lacking domain-specific knowledge to deepen the understanding of entities and other nouns~\citep{DBLP:conf/konvens/OstendorffBBSRG19} (which are often the answers in span-extraction MRC tasks). Without proper descriptions of such entities in the passage, MRC models often fail to understand and extract key information~\citep{DBLP:conf/iclr/DasMYTM19}.
Hence, the explicit fusion of knowledge in MRC models is vital for learning context-aware token representations ~\citep{DBLP:conf/acl-mrqa/PanSYCJCY19, DBLP:conf/emnlp/QiuZFLJLLZ19, DBLP:journals/corr/abs-2008-02434}. Instead of encoding entities appearing in both knowledge bases and passages into the MRC model only ~\citep{ DBLP:conf/acl/InkpenZLCW18}, our proposed model encodes all the triples from a medical KG and then employs heuristic rules to retrieve relevant entities. This practice allows the model to acquire deeper understanding of domain-specific terms.

\section{The CMedMRC Dataset}
In this section, we briefly describe the collection process and provide an  analysis on various aspects of the CMedMRC dataset. For more dataset collection and statistical analysis of dataset details, we refer readers to the Appendix \ref{appendix_A} and Appendix \ref{appendix_B}.
\begin{table*}
\tiny
\centering
\begin{tabular}{clc}
\toprule
\textbf{Skills}  & \textbf{Example}  & \textbf{Percentage}  \\ \midrule

\textbf{Token matching} & \makecell*[l]{\textbf{\textit{Passage}}: \tiny...急性羊水过多较少，见多发生在孕20～24周，羊水急剧增多，子宫短期内明显增大...\\(...it is less likely to secrete too much acute \underline{\color{blue} \textit{amniotic fluid}}. The disease is most common in the 20 to 24 weeks of \underline{\color{blue} \textit{pregnancy}}. \\The amniotic fluid increases sharply with the uterus enlarged significantly in the short term...)\\\textbf{\textit{Question}}: \tiny 怀孕期间羊水什么时候分泌过多? (When does the \underline{\color{blue} \textit{amniotic fluid}} secrete too much during \underline{\color{blue} \textit{pregnancy}}?) \\ \textbf{\textit{Answer}}: 20～24周 (20～24 weeks) }  & 31\%  \\ \midrule

\textbf{Co-reference resolution} & \makecell*[l]{\textbf{\textit{Passage}}: \tiny...尖锐湿疣有「割韭菜」的臭名声，它的治疗瓶颈在于病毒不进入血循环，因此机体无法产生免疫应答，\\所以容易反复复发。...(...\underline{\color{blue} \textit{genital warts}} has a bad reputation of cutting leeks. The bottleneck of \underline{\color{blue} \textit{its}} treatment is that the virus \\does not enter the blood circulation, so the body cannot produce an immune response and \underline{\color{blue} \textit{it}} is easy to relapse repeatedly....)\\\textbf{\textit{Question}}: \tiny 为什么尖锐湿疣易反复
发作? (Why \underline{\color{blue} \textit{genital warts}} is easy to relapse repeatedly?) \\ \textbf{\textit{Answer}}: 病毒不进入血循环，因此机体无法产生免疫应答 (The virus does not enter the blood circulation, so the body\\ cannot produce an immune response) }  & 11\%  \\ \midrule

\begin{tabular}[c]{@{}c@{}}\textbf{Multi-sentence}\\ \textbf{reasoning}\end{tabular} &\makecell*[l]{\textbf{\textit{Passage}}: \tiny... 老年人应在医师指导下使用。5.肝、肾功能不全者慎用。6.孕妇及哺乳期妇女慎用。...\\(...\underline{\color{blue} \textit{The elderly should take the medicine under the guidance of a physician. 5. Use with caution in patients with liver and }} \\\underline{\color{blue} \textit{kidney insufficiency. 6. Use with caution in pregnant and lactating women.}})\\\textbf{\textit{Question}}: \tiny 哪些人群慎用此药品? (Which groups of people should use this drug with caution?)\\\textbf{\textit{Answer}}:\tiny 老年人应在医师指导下使用。5.肝、肾功能不全者慎用。6.孕妇及哺乳期妇女慎用\\(The elderly should take the medicine under the guidance of a physician. 5. Use with caution in patients with liver and \\kidney insufficiency. 6. Use with caution in pregnant and lactating women.)} & 18\% \\ \midrule

\begin{tabular}[c]{@{}c@{}}\textbf{Implicit causality} \end{tabular} &\makecell*[l]{\textbf{\textit{Passage}}: \tiny... 不是所有的白细胞减少都必须治疗的，
关键看白细胞减少的程度、机体的一般状态以及医生的建议； \\...... 因为无症状的白细胞减少对生活的影响是很小的；...\\(...\underline{\color{blue} \textit{Not all leukopenia must be treated.}} The key depends on the degree of leukopenia, the general state of the body and the \\doctor's advice;......\underline{\color{blue} \textit{Because asymptomatic leukopenia has little impact on life}}...)\\\textbf{\textit{Question}}: \tiny 为什么不是所有的白细胞减少都要进行治疗? (Why do \underline{\color{blue} \textit{not all leukopenia have to be treated?}})\\\textbf{\textit{Answer}}:\tiny 因为无症状的白细胞减少对生活的影响是很小的 (Because asymptomatic leukopenia has little impact on life.)} & 22\% \\ \midrule

\textbf{Domain knowledge} & \makecell*[l]{\textbf{\textit{Passage}}:\tiny ...发病率居遗传性血小板功能缺陷疾病的首位。血栓细胞衰弱发病多见于幼年，发病率为 0.01/万...\\(...The incidence is the highest in hereditary platelet dysfunction diseases. 
The incidence of \underline{\color{blue} \textit{thrombotic cell weakness}} is more \\ common in childhood with an incidence rate of 0.01 / 10,000...)\\\textbf{\textit{Question}}:\tiny 血小板无力症的发病率约为多少？
\\(What is the incidence rate of \underline{\color{blue} \textit{blood platelet weakness?}} )\\\textbf{\textit{Answer}}: \tiny 0.01/万 (0.01/10,000)} & 16\% \\ \bottomrule
\end{tabular}
\caption{\label{answer_question_methods} Reading comprehension skills of models required to answer questions in CMedMRC.
The blue and underscore contents in brackets indicate why the sample belongs to its category. (Best viewed in color)}
\end{table*}

\subsection{Dataset Collection Process}
The dataset collection process follows the SQuAD-style ~\citep{DBLP:conf/emnlp/RajpurkarZLL16} rather than collecting question-answer pairs as in Google Natural Questions~\citep{DBLP:journals/tacl/KwiatkowskiPRCP19}.
Our medical text corpus is collected from DXY Medical~\footnote{\url{http://www.dxy.cn/}}, an authoritative medical knowledge source in China.
The general data collection process of CMedMRC consists of four major steps: passage collection, question-answer pair collection, support sentence selection and additional answer construction. 
Briefly speaking, during the passage collection process, we filter the corpus to generate high-quality medical passages.
A group of human annotators are required to ask questions on medical knowledge and annotate the answers from these passages. The annotation results are in the form of question-answer pairs.
Following SQuAD, we ask annotators to provide 2 additional answers for each question in the DEV and TEST sets.


Since people are concerned about the scientific explanation and sources of answers in the medical domain, we ask annotators to select the support sentence of their annotated answer similar to those of CoQA ~\citep{DBLP:journals/tacl/ReddyCM19} and QuAC~\citep{DBLP:conf/emnlp/ChoiHIYYCLZ18}. 
Finally, CMedMRC consists of three parts: 12,700 training samples, 3,630 development samples and 1,823 testing samples.

\subsection{Quality Control}
During the dataset collection process, we take the following measures to ensure the quality of the dataset. i) The knowledge source (DXY Medical) contains high-quality medical articles which are written by medical personnel and organized based on different topics in the medical domain.
ii) Our annotators are all engaged in medical-related professions rather than annotators with short-term guidance only.
iii) We further hire 12 medical experts to check all the collected samples rather than checking a randomly selected sample only.
The experts remove out-of-domain questions and questions that are unhelpful to medical practice. In this stage, the experts are divided into two groups and cross-check their judgments.

\subsection{Challenges of Understanding Texts}
Due to the closed-domain property of our dataset, there are some domain-specific textual features in both passages and questions that the model needs to understand.
Based on our observations of the CMedMRC, we summarize the following two major challenges.
These challenges can be also regarded as key reasons why some recent state-of-the-art MRC models cannot address the medical MRC task on CMedMRC well.

\textit{Lexical-Level:} i) Long-tail terminology means these medical terms occur very infrequently and are prone to Out-Of-Vocabulary (OOV) problems. ii) Synonym terminology means that some medical terms may express the same meaning, but there is a distinction between colloquial expressions and professional terms. The above two points require the model to have rich domain knowledge to solve. iii) Terminology combination means these terms are usually formed by a combination of multiple terms, while one term is the attributive of another. This  
   does not only require the model to have domain knowledge but also poses challenges to phrase segmentation in specific domains.
   
\textit{Sentence-Level:} Paraphrasing means some words in questions are semantically related to certain tokens in passages, but are expressed differently. Consider the last question in Table~\ref{question_challenge_type}. When the model tries to answer the ``not-recommended'' question, it should focus on negative terms (``cause infection'' and ``delay the healing of the wound'').

\subsection{Reasoning Skills for MRC Models}
We randomly select 100 samples from the development set to analyze what skills the model should have in order to answer the questions correctly.
We divide the reasoning skills corresponding to these samples into five major categories, namely token matching, co-reference resolution, multi-sentence reasoning, implicit causality and domain knowledge.
Examples are shown in Table~\ref{answer_question_methods}.
It is particularly noteworthy that the fifth type is the need of domain knowledge to answer medical questions.
Consider the example:


\textit{Passage: ... The incidence of thrombotic cell weakness is most common in childhood with an incidence rate of 0.01 / 10,000...}

\textit{Question: What is the incidence rate of blood platelet weakness?}

\textit{Answer: 0.01/10,000.}

We know that the \textit{blood platelet} in the question refers to the \textit{thrombotic cell} described in the passage through the medical knowledge base.
It shows that the rich information of the knowledge base can help the model obtain a better understanding of domain terms to improve the MRC performance.

\section{The CMedBERT Model}

\subsection{Task Formulation and Model Overview}
\begin{figure*}
\centering
\includegraphics[height=6cm]{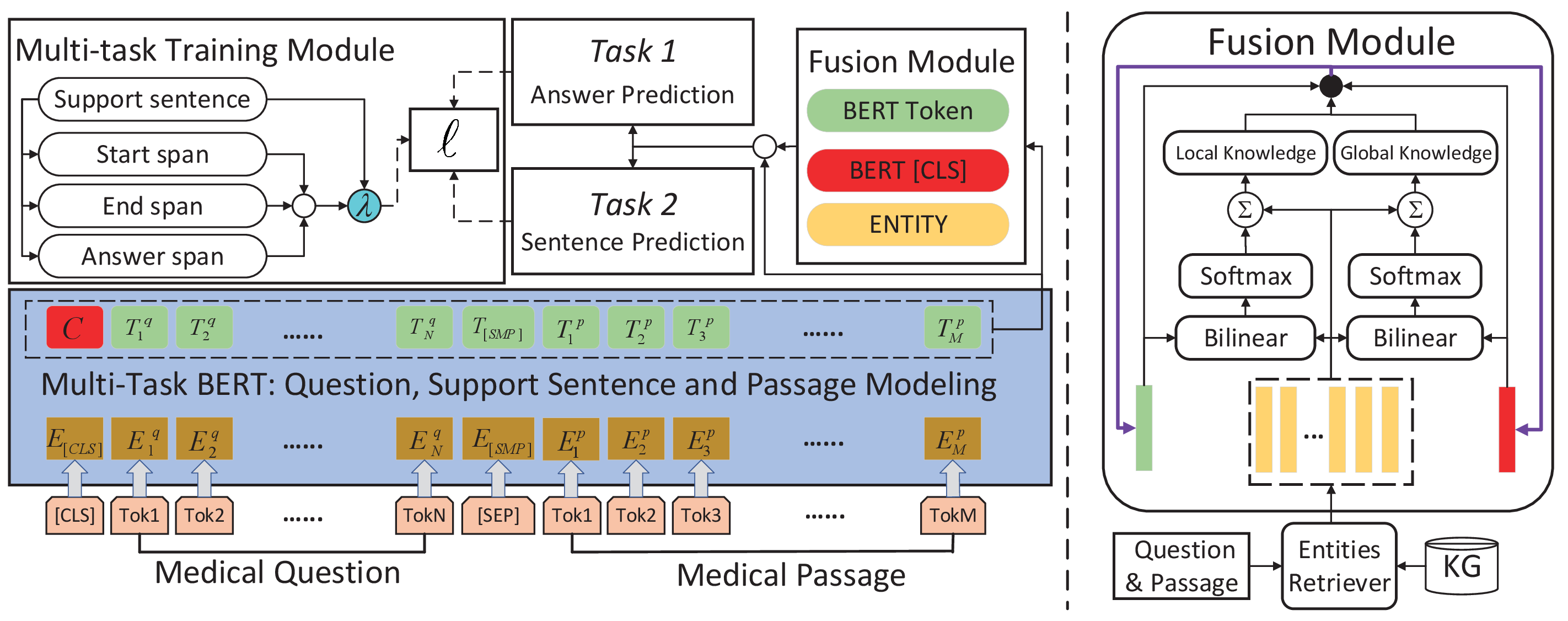}
\caption{Model overview. The green box and the red box in the heterogeneous feature fusion layer represent the local and global token information, respectively. In the multi-task training module, the model learns the relationship between two tasks by dynamically learning the parameter $\lambda$. (Best viewed in color)}
\label{CMedBERT_model}
\end{figure*}

For our task, the input includes a medical question $Q$ together with the passage $P$. Let $\left\{ p_1, p_2,\cdots p_m \right\}$ and $\left\{ q_1, q_2,\cdots q_n \right\}$ represent the passage and question tokens, respectively.
In the answer prediction task, the goal is to train an MRC model which extracts the answer span $\left\{ p_i, p_{i+1},\cdots p_j \right\}$ ($0 \leq i \leq j \leq m$) from $P$ that correctly answers the question $Q$. Additionally, the model is required to predict the support sentence tokens $\left\{ p_k, p_{k+1},\cdots p_l \right\}$ ($0 \leq k \leq l \leq m$) from $P$ to provide additional medical knowledge and to enhance interpretability of the extracted answers. We constrain that $\left\{ p_k, p_{k+1},\cdots p_l \right\}$ must form a complete sentence, instead of incomplete semantic units and the support sentence tokens contain the answer span.
The high-level architecture of the CMedBERT model is shown in Figure~\ref{CMedBERT_model}. It mainly includes four modules: BERT encoding, knowledge embedding and retrieval, heterogeneous feature fusion and multi-task training.

\subsection{BERT Encoding}
This module is used to learn context-aware representations of question and the passage tokens.
For each input pair (the question $Q$ and the passage $P$), we treat $[\langle CLS \rangle, Q, \langle SEP \rangle, P, \langle SEP \rangle ]$ as the input sequences for BERT.
We denote $\left\{ h_i \right\}_{i=1}^{m+n+3}$ as the hidden layer representations of tokens, where $m$ and $n$ are the length of passage tokens and question tokens, respectively.

\subsection{Knowledge Embedding and Retrieval}
\label{Knowledge_Embedding_Retrieval}
In the knowledge bases, relational knowledge is stored in the form of (\textit{subject, relation, object}) triples.
In order to fuse knowledge into token representations, we first encode all entities in the knowledge base into a low-dimensional vector space.
Here, we employ PTransE~\cite{DBLP:conf/emnlp/LinLLSRL15} to learn entity representations, and denote the underlying entity embedding as $e_i$.
Because existing medical NER tools do not have high coverage over our corpus, we consider five types of token strings as candidate entities: noun, time, location, direction and numeric. Two matching strategies are then employed to retrieve relevant entities from the knowledge base:
(\romannumeral1) The two strings match exactly. 
(\romannumeral2) The number of overlapped tokens is larger than a threshold. After relevant entities are retrieved, we can fuse the knowledge into contextual representations, introduced below.

\subsection{Heterogeneous Feature Fusion}

In this module, we fuse heterogeneous entity features retrieved from the knowledge base into the question and passage tokens representations.

\textbf{Local Fusion Attention.} We observe that each token is usually related to multiple entities of varying importance.
Thus, we assign different weights to the entity embedding $e_j$ corresponding to the token representation $h_{i}$ using attention mechanism:
\begin{equation}
    \alpha_{i,j} = \frac{exp(e_j^{T}Wh_{i})}{\sum\nolimits_{k=1}^{K}{exp(e_k^{T}Wh_{i})}}
\end{equation}
where $K$ is the number of entities and $\alpha_{i,j}$ represents the similarity between the $j^{th}$ entity in the retrieved entity set and the $i^{th}$ token. $W \in \mathbb{R}^{d_2 \times d_1}$ where  \textit{${d_1}$} is the dimension of BERT's output and \textit{$d_2$} is the dimension of entity embeddings. After fusing, the representation of the $i^{th}$ token is:
$\bar{e_i} = \sum\nolimits_{k=1}^{K}{\alpha_{i,k}e_k}$.
However, $\bar{e_i}$ is only related to retrieved entities, not other tokens in the question-passage pair.

\textbf{Global Fusion Attention.} In BERT, the output of the $[CLS]$ tag represents the entire sequence information learned by transformer encoders.
We use the token output $h_{[CLS]}$ to model the knowledge fusion representation of the entire entity collection that each token recalls:
\begin{gather}
    \beta_{[CLS],j} = \frac{exp(e_j^{T}Wh_{[CLS]})}{\sum\nolimits_{k=1}^{K}{exp(e_k^{T}Wh_{[CLS]})}} \\
    \hat{e_i} = \sum\nolimits_{k=1}^{K}{\beta_{[CLS],k}e_k}
\end{gather}
where $\hat{e_i}$ is the global knowledge fusion result  corresponding to the $i^{th}$ token.

\textbf{Gated Loop Layer.}
In order to fuse local and global results into token representations, we design a gated loop layer.
The information of knowledge fusion is filtered through the gating mechanism in each loop of modeling. In the initialization stage, we simply have $h_i^0=h_i$. In the $l^{th}$ iteration, we have the following update process:
\begin{gather}
    G_{i}^\ell = \sigma (tanh(W[h_{i}^\ell, \bar{e_{i}}, \hat{e_{i}}])) \\
    h_i^{\ell+1} =  G_{i}^\ell\odot h_{i}^\ell
\end{gather}
This process runs for $L$ loops and this fusion process output is $h_i^{L}$.
The loop process mimics the human's behavior of reading the passage repeatedly to find the most accurate answers.

\begin{table*}[]
\small
\centering
\begin{tabular}{lccccccccc}
\toprule
  \multirow{3}*{Model}& \multicolumn{4}{c}{Answer} & \multicolumn{4}{c}{Support Sentence} \\ 
\cmidrule(r){2-5} \cmidrule(r){6-9}
 & \multicolumn{2}{c}{Exact Match (EM)} & \multicolumn{2}{c}{F1} & \multicolumn{2}{c}{Exact Match (EM)}  & \multicolumn{2}{c}{F1} \\
\cmidrule(r){2-3} \cmidrule(r){4-5}\cmidrule(r){6-7}\cmidrule(r){8-9}
 & Dev & Test & Dev & Test & Dev & Test & Dev & Test \\
\midrule
DrQA & 42.00\% & 37.45\% & 58.66\% & 57.15\% & 5.07\% & 5.88\% & 30.24\% & 32.52\% \\
BERT\_base & 64.83\% & 68.31\% & 81.08\% & 83.74\%  & 21.42\% & 17.70\% & 52.27\% & 48.34\% \\
ERNIE & 65.49\% & 68.57\% & 81.17\% & 83.86\% & 20.71\% & 16.32\% & 49.88\% & 45.81\% \\
KT-NET$^\spadesuit$ & 64.58\% & 69.03\% & 81.06\% & 84.18\%  & 15.42\% & 13.48\%  & 49.37\% & 46.45\%  \\
MC-BERT & 66.58\% & 68.62\% & 81.23\% & 83.98\%  & 20.08\% & 16.77\% & 47.33\% & 44.53\% \\
KMQA & 66.19\% & 68.45\% & 81.20\% & 83.79\%  & 20.63\% & 16.54\% & 49.27\% & 45.97\% \\
\midrule
CMedBERT$^\clubsuit$ &  69.00\% &  72.84\% &  82.68\% &  85.38\%  &  25.17\% &  24.18\% &  52.36\% & 49.69\% \\ 
CMedBERT$^\spadesuit$ & \bf 70.33\% & \bf 72.91\% & \bf 83.43\% & \bf 85.64\%  & \bf 25.58\% & \bf 25.51\% & \bf 52.67\% & \bf 52.41\% \\
\bottomrule
\end{tabular}
\caption{The results of multi-task prediction (answer and support sentence) over CMedMRC.}
\label{multi_model_perfermance}
\end{table*}

\subsection{Multi-task Training}
The output layer of CMedBERT is extended from BERT.
We first concatenate two types of token representations and calculate the probability of the $i^{th}$ token being selected in the support sentence as follows:
\begin{gather}
    o_{i} = \sigma (W[h_{i}, h_i^{L}]), \ p^{support}_{i} = \sigma (Wo_{i})
\end{gather}
We also calculate its probabilities as the starting and the ending positions of the answer span, respectively:
\begin{align*}
     p_i^{start}=\frac{exp(w_1^To_i)}{\sum_{j}{exp(w_1^To_j)}},\  p_i^{end}=\frac{exp(w_2^To_i)}{\sum_{j}{exp(w_2^To_j)}}
\end{align*}

The loss function of the answer prediction task is the negative log-likelihood of the starting and ending positions of ground-truth answer tokens:
\begin{gather}
    \mathcal{L_{A}} = -\frac{1}{N}\sum_{j=1}^{N}(logp_{y_j^{start}}^{start}+logp_{y_j^{end}}^{end})
\end{gather}
For the extraction of support sentences, the loss function is defined by cross-entropy:
\begin{gather}
    \mathcal{L_{S}} = -\frac{1}{N}\sum_{j=1}^{N}\sum_{i=1}^{M}(y^{support}_{j}log p^{support}_{i})
\end{gather}
where $N$ is the number of samples and $M$ is the length of input sequences.
$y_j^{start}$,$y_j^{end}$ is the starting and ending positions of ground-truth of the $j^{th}$ token.
Furthermore, if the token is in the support sentence, the token label $y^{support}_{j}$ is set to 1, and 0 otherwise.

The representations of the support sentence are related to the positions of the answer.
In order to better model the relationship between two tasks, we dynamically learn the coefficient between the loss values of two tasks. 
Let $h_{su}$ and $o_{sp}$ be self-attended, averaged pooled representations of the support sentence and the answer span. $o_{st}, o_{ed}$ are the start and end position token representations of the answer, respectively. We have:
\begin{gather}
    \gamma_{st}, \gamma_{ed}, \gamma_{sp} = h_{su}[o_{st}, o_{ed}, o_{sp}]^T \\
    H_{A} = \sigma  (W[\gamma_{st}o_{st}, \gamma_{ed}o_{ed},  \gamma_{sp}o_{sp}])
\end{gather}
where $\gamma_{st}, \gamma_{ed}, \gamma_{sp}$ are the weight coefficients between the supporting sentence and the start/end/total token representations of the answer span.
The loss value coefficient of two tasks $\lambda$ and the total loss $\mathcal{L}$ are as follows:
\begin{gather}
    \lambda = max\{0, \cos(H_{A}, h_{su})\} \\
    \mathcal{L} =  \mathcal{L_{A}} + \lambda\mathcal{L_{S}}
\end{gather}
We minimize the total loss $\mathcal{L}$ to update our model parameters in the training process.


\section{Experiments and Result Analysis}
\subsection{Experimental Setups}
We evaluate CMedBERT on CMedMRC, and compare it against six strong baselines: DrQA \citep{DBLP:conf/acl/ChenFWB17}, BERT\_base \citep{DBLP:conf/naacl/DevlinCLT19}, ERNIE \citep{DBLP:conf/acl/ZhangHLJSL19}, KT-NET \citep{DBLP:conf/acl/YangWLLLWSL19}, MC-BERT~\citep{DBLP:journals/corr/abs-2008-10813} and KMQA \citep{DBLP:conf/emnlp/LiHCPW20}.
KT-NET is the first model to leverage rich knowledge to enhance pre-trained language models for MRC.
MC-BERT is the first Chinese biomedical pre-trained model fine-tuned on BERT\_base.
We only use the encoder layer in KMQA removing the answer layer due to the different answer type.

For evaluation, we use EM (Exact Match) and F1 metrics for answer and support sentence tasks. We calculate character-level overlaps between prediction and ground truth for the Chinese language, rather than token-level overlaps for English. To assess the difficulty of solving CMedMRC tasks, we select 100 testing samples to evaluate human performance.
Human scores of EM and F1 are 85.00\% and 96.69\% for answer prediction, respectively.

In the implementation, we set the learning rate as
5e-5 and the batch size as 16, and the max sequence length as 512.
Other BERT's hyper-parameters are the same as in Google's settings~\footnote{\url{https://github.com/google-research/bert}}. Each model is trained for 2 epochs by the Adam optimizer \citep{DBLP:journals/corr/KingmaB14}. 
Results are presented in average with 5 random runs with different random seeds.
Other implementation details are in Appendix \ref{experimental_result}.

\begin{table}[]
\small
\centering
\begin{tabular}{lccccc}
\toprule
  \multirow{2}*{Model}& \multicolumn{2}{c}{Exact Match (EM)} & \multicolumn{2}{c}{F1} \\ 
\cmidrule(r){2-3} \cmidrule(r){4-5}
 & Dev & Test & Dev & Test\\  
\midrule
DrQA & 34.50\% & 32.10\% & 56.67\% & 56.64\%\\
BERT\_base & 62.39\% & 68.29\% & 81.48\% & 83.70\% \\
ERNIE & 63.18\% & 66.92\% & 81.74\% & 83.41\% \\
KT-NET$^\spadesuit$ & 64.64\% & 66.26\% & 82.48\% & 83.61\% \\
MC-BERT & 63.39\% & 68.38\% & 81.86\% & 83.88\% \\
KMQA & 64.37\% & 67.48\% & 81.95\% & 83.74\% \\
\midrule
CMedBERT$^\clubsuit$ & 68.00\% & 72.11\% & 82.50\% & 85.33\% \\
CMedBERT$^\spadesuit$ & \textbf{69.83\%} & \textbf{72.84\%} & \textbf{83.02}\% & \textbf{85.54\%} \\
\midrule
Human & - & 85.00\% & - & 96.69\%  \\ \bottomrule
\end{tabular}
\caption{Result of single-task (answer prediction). $^\clubsuit$ and $^\spadesuit$ indicate that CMedBERT uses BERT\_base and MC-BERT as the encoder, respectively.}
\label{single_model_perfermance}
\end{table}

\subsection{Model Results}
Table \ref{multi_model_perfermance} and Table \ref{single_model_perfermance} show the multi-task and single-task results on the CMedMRC development and testing sets. CMedBERT has a great improvement compared to four strong baseline models in both tasks.
Specifically, our CMedBERT outperforms the state-of-the-art model by a large margin in multi-task results, with \textbf{+3.88\%EM / +1.46\%F1} improvements, which shows the effectiveness of our model.
Meanwhile, in the support sentence task, our model also has the best performance, improving (\textbf{+7.81\%EM / +4.07\%F1}) over the testing set.
In single task evaluation, we remove the support sentence training module and the dynamic parameter for loss function module.
Our model improves (\textbf{+4.46\%EM / +1.66\%F1}) over the best baseline model. 
In addition, we find that using support sentence prediction as an auxiliary task and the pre-training technique in medical domain can further improve the performance of CMedBERT.

\begin{table}[]
\small
\centering
\setlength{\tabcolsep}{1.5mm}{
\begin{tabular}{ccccc}
\toprule
  \multirow{2}*{Model} & \multicolumn{2}{c}{Answer} & \multicolumn{2}{c}{Sentence} \\ 
  \cmidrule(r){2-5}
 & EM & F1 & EM & F1\\ \midrule
CMedBERT$^\spadesuit$ & \bf 72.91\% & \bf 85.64\% & \bf 25.51\% & \bf 52.41\% \\ \midrule
w/o Local Att. & 68.93\% & 83.89\% & 19.75\% & 49.45\% \\
w/o Global Att. & 71.71\% & 84.96\% & 17.59\% & 47.01\%\\
w/o $\lambda$ & 71.91\% & 85.09\% & 21.09\% & 48.80\% \\ \bottomrule
\end{tabular}}
\caption{Ablation study of CMedBERT (testing set).}
\label{table_ablation_study}
\end{table}

\subsection{Ablation study}
In Table \ref{table_ablation_study}, we choose three important model components for our ablation study and report the results over the testing set.
When the dynamic parameter $\lambda$ of the loss function is removed from the model, the performance of the model on two tasks is decreased by (\textbf{-1.00\%EM and -0.55\%F1}) and (\textbf{-4.42\%EM and -3.61\%F1}), respectively.
Without local attention, the EM performance in the answer prediction task decreases by (\textbf{-3.98\%EM and -1.75\%F1}).
Experiments have shown that the model performs worse without the local fusion attention than without the global fusion attention and the dynamic parameter $\lambda$.
However, the performance of support sentence task is decreased significantly by (\textbf{-7.92\%EM and -3.61\%F1}) without global fusion attention.
It shows that local fusion attention is more important for extracting answer spans, while global fusion attention plays a larger role in support sentence prediction.

\begin{figure}[htbp]

    \subfigure[Tokens probabilities of \textbf{CMedBERT$^\spadesuit$}]{
        \begin{minipage}[t]{0.5\textwidth}
            \centering
            \includegraphics[width=7cm]{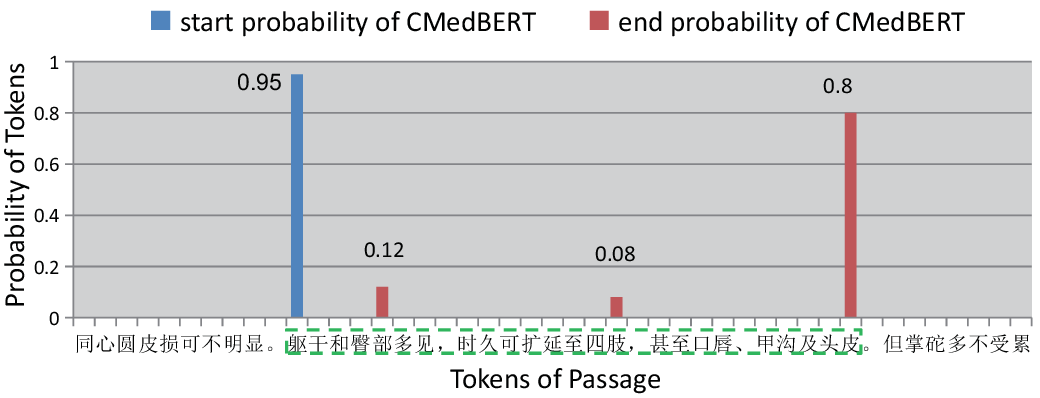}
        \end{minipage}
    }
    
    \subfigure[Tokens probabilities of \textbf{BERT\_base}]{
        \begin{minipage}[t]{0.5\textwidth}
            \centering
            \includegraphics[width=7cm]{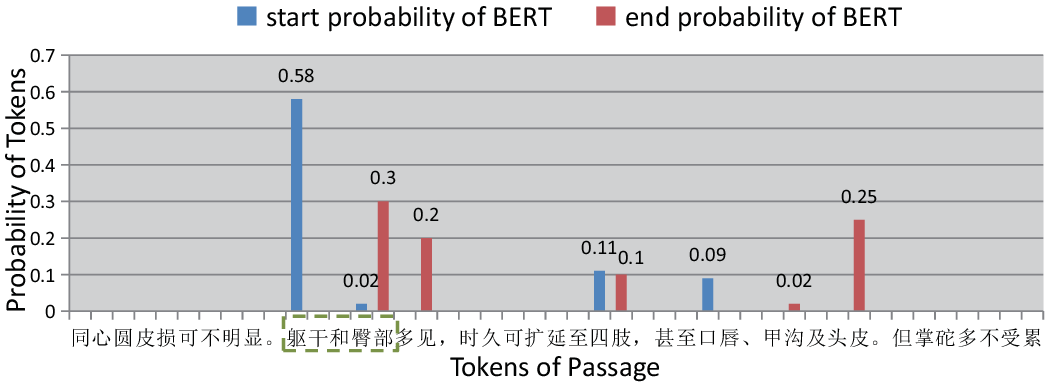}
        \end{minipage}
    }

    \caption{Case study. Predicted answer spans are in the green dotted box. Product of the maximum starting and ending probabilities of CMedBERT is \textbf{0.76}, with BERT to be \textbf{0.174}.}
    \label{start_end_prob}
\end{figure}

\subsection{Case Study}
In Figure~\ref{start_end_prob}, we use our motivation example to conduct a case study.
In BERT, we can see that the difference among the probability values of different words is small, especially when predicting the probability of ending positions.
The ending position probabilities of token ``部" token ``皮" are \textbf{0.3} and \textbf{0.25}, leading the model to extract the wrong answer span.
However, in the knowledge retrieval module of CMedBERT, multiple entities representation are fused into the context-aware latent space representation to enhance the medical text semantic understanding.
Therefore, in our CMedBERT model, the starting position probability is \textbf{0.95} and the end position probability is \textbf{0.8}.
In this case, the CMedBERT model can easily choose the correct range of the answer span.

\subsection{Discussion of Support Sentence Task}
Compared with the answer prediction task,  existing models have poor prediction results on the EM metric in the support-sentence task.
In prediction results, we randomly select 100 samples for analysis.
We divide the error types into the following three main types (see Appendix): i) starting position cross ii) ending position cross iii) answer substring.
The most common error type is the answer substring, accounting for \textbf{46\%}.
In this error type, the predicted result of our model is part of the true result, which shows the model cannot predict long answers completely~\citep{DBLP:conf/acl/YuanSBGLDFJ20} and reduce the accuracy of the results greatly.

\section{Conclusion}
In this work, we address medical MRC with a new dataset \textbf{CMedMRC} constructed. An in-depth analysis of the dataset is conducted, including statistics, characteristics, required MRC skills, etc.
Moreover, we propose the \textbf{CMedBERT} model, which can help the pre-trained model better understand domain terms by retrieving entities from medical knowledge bases. Experimental results confirm the effectiveness of our model. In the future, we will explore how knowledge can improve the performance of models in the medical domain.

\section*{Acknowledgements}
We would like to thank anonymous reviewers for their valuable comments. 
This work is supported by the National Key Research and Development Program of China under Grant No. 2016YFB1000904, and Alibaba Group through Alibaba Research Intern Program.

\bibliographystyle{acl_natbib}
\bibliography{anthology,acl2021}
\clearpage
\appendix
\section{Detailed Dataset Construction Process of CMedMRC}
\label{appendix_A}
\subsection{Medical Passage Curation}
We use the following rules to obtain 20,000 passages from DXY as the inputs to human annotators: 
\begin{itemize}
\item We use regular expressions to filter out images, tables, hyperlinks, etc. The English-Chinese translations of medical terms are also provided if the passages contain medical terms in English.


\item We find that if we follow \citep{DBLP:conf/emnlp/RajpurkarZLL16} to limit the lengths of the passages within 500 tokens, our human annotators could not ask 4 high-quality medical questions easily. Hence,  our passage length limit is 1000 tokens.
\end{itemize}

\subsection{Medical Question-Answer Pair Collection}
We employ a group of annotators with professional medical background to generate question-answer pairs from the medical passages. Here are some general guidelines:
\begin{itemize}
\item We encourage our annotators to ask questions to which the answers are uniformly distributed in different positions of medical passages.
\item For each medical passage, we limit the number of questions to 4.
\item Each question should be strictly related to the medical domain. When creating the questions, no any part of the texts can be directly copied and pasted from the given medical passages.
\item We limit the number of answer tokens to no more than 40.
\end{itemize}

\begin{table}[t]
\footnotesize
\centering
\begin{tabular}{@{}clc@{}}
\toprule
\textbf{Answer Type} & \textbf{Pct.} & \textbf{Example} \\ \midrule
Numeric &\makecell*[c]{6\%} & 20\%  \\ \midrule
Time/Date & \makecell*[c]{11\%} & \makecell*[c]{1-3小时 (1-3 Hours)} \\ \midrule
Person & \makecell*[c]{8\%}  & \makecell*[c]{儿童 (Child)} \\ \midrule
Location & \makecell*[c]{5\%}  &\makecell*[c]{安徽, 云南, 湖北\\(Anhui, Yunnan, Hubei)} \\ \midrule
\makecell*[c]{Noun Phrase}  & \makecell*[c]{18\%}  & \makecell*[c]{输卵管炎 (Salpingitis)} \\ \midrule
Verb Phrase &\makecell*[c]{6\%} & \makecell*[c]{清洗，干燥和粉碎\\(Wash, dry and crush)} \\ \midrule
Yes/No &\makecell*[c]{1\%}  &\makecell*[c]{不会感染 (Will not infect)} \\ \midrule
Description & \makecell*[c]{44\%}  & \makecell*[c]{维生素缺乏 (Vitamin deficiency)} \\ \midrule
Other & \makecell*[c]{1\%}  & \makecell*[c]{严重 (Severe)} \\  \bottomrule
\end{tabular}
\caption{\label{answer_type_table} Statistical results for answer types.}
\end{table}

\begin{table}[t]
\small
\centering
\begin{tabular}{cccc}
\toprule
 & Train & Dev & Test \\ \midrule
\# Questions & 12,700 & 3,630 & 1,823  \\
Avg. tokens of passages & 883.64  & 743.10 & 745.52 \\
Avg. tokens of questions & 15.40  & 14.85 & 15.23 \\
Avg. tokens of answers & 19.69  & 18.48 & 16.57 \\
Avg. tokens of support sen. & 57.50  & 48.19 & 42.70 \\ \bottomrule
\end{tabular}
\caption{\label{average_length_data} Statistical results of text length in our CMedMRC dataset.}
\end{table}

\subsection{Support Sentence Selection}
In our dataset, we add an index to each sentence in the passages. Annotators are required to select the support sentence index and mark the range of the answer spans on the user interface. 

\subsection{Additional Answer Construction}
To evaluate the human performance of our dataset and make our model more robust, we collect two additional answers for each question in the development and testing sets. We employ another 12 annotators for answer construction. Since our medical passage is relatively long, we show the questions and the passage contents again on the interface, together with the previously labeled support sentence indices.

\begin{table*}
\tiny
\centering
\begin{tabular}{clc}
\toprule
\textbf{Error Type}  & \textbf{Example}  & \textbf{Percentage}  \\ \midrule

\textbf{Start position cross} & \makecell*[l]{\textbf{\textit{Ground-truth}}: \tiny当有急性炎症或者化脓时，会有剧烈疼痛；或者合并牙神经发炎时也会出现剧烈疼痛。\\({\color{blue} \textit{When there is acute inflammation or suppuration, there will be severe pain;}} or when combined with \\ dental nerve inflammation, there will also be severe pain) \\\textbf{\textit{Prediction}}: \tiny 有轻微的隐痛或胀痛；当有急性炎症或者化脓时，会有剧烈疼痛； \\ (There is slight dull pain or pain; {\color{blue} \textit{when there is acute inflammation or suppuration, there will be severe pain;}})}  & 25\%  \\ \midrule

\textbf{End position cross} & \makecell*[l]{\textbf{\textit{Ground-truth}}: \tiny以下人群高危：乙肝、丙肝病毒慢性感染者；患有类风湿关节、狼疮、硬皮病等免疫\\ 性疾病；吸烟。 ......遗传因素对本病起到一定作用。\\(The following people are at high risk: people with chronic hepatitis B and C virus infections; suffering from\\ immune diseases such as rheumatoid joints, lupus, and scleroderma; smoking.
...... {\color{blue} \textit{Genetic factors play a role in this disease.}})\\\textbf{\textit{Prediction}}: \tiny 目前认为血管炎是一种自身免疫性疾病，......
遗传因素对本病起到一定作用。一些药物\\也可以引起血管炎，还有一些感染（如丙肝、乙肝、梅毒）也可以引起血管炎。 \\ ( {\color{blue} \textit{At present, vasculitis is considered to be an autoimmune disease, ......
Genetic factors play a role in this disease.}}
\\ Some drugs can also cause vasculitis, and some infections (such as hepatitis C, hepatitis B, and syphilis) can also cause vasculitis.) }  & 21\%  \\ \midrule

\textbf{Answer substring} & \makecell*[l]{\textbf{\textit{Ground-truth}}:\tiny 这些药物具有抗炎、改善毛细血管通透性、减轻水肿、止痛等作用，同时对日光性\\皮炎有很好的治疗作用。\\(These drugs have {\color{blue} \textit{anti-inflammatory, improve capillary permeability, reduce edema, pain relief, etc.}}, and have a good therapeutic\\ effect on solar dermatitis.)\\\textbf{\textit{Prediction}}:\tiny 具有抗炎、改善毛细血管通透性、减轻水肿、止痛等作用
\\({\color{blue} \textit{Anti-inflammatory, improve capillary permeability, reduce edema, relieve pain, etc.}})} & 46\% \\ \midrule

\textbf{Other} & \makecell*[l]{\textbf{\textit{Ground-truth}}:\tiny 抗组胺药第一代的经典代表药「马来酸氯苯那敏」就是一个，它俗称扑尔敏，在多年临床应用中\\没有发现对胎儿有明显的致畸或其他严重危害。\\(One of the classic representative drugs of the first generation of antihistamines is "Chlorpheniramine Maleate". It is commonly \\known as Chlorpheniramine. It has not been found to have obvious teratogenic or other serious harm to the fetus in many years \\of clinical application.) \\\textbf{\textit{Prediction}}:\tiny 但临床上也有一些药物是经过多年验证，只要注意把握用药时间和药量，即使让孕妇吃也不会有事的
\\(However, there are also some drugs that have been verified for many years in clinical practice. As long as you pay attention \\ to the time and amount of medication, it will be fine even if pregnant women take it.)} & 8\% \\ \bottomrule
\end{tabular}
\caption{\label{support_sentence_answer} Three typical error answer types in support sentence task. 
The blue and underscore contents in brackets indicate why the sample belongs to its corresponding category.} 
\end{table*}

\section{Statistical Analysis of the CMedMRC Dataset}
\label{appendix_B}
\subsection{Question and Answer Types}
\begin{figure}[t]
\centering
\includegraphics[height=5cm, width=7.5cm]{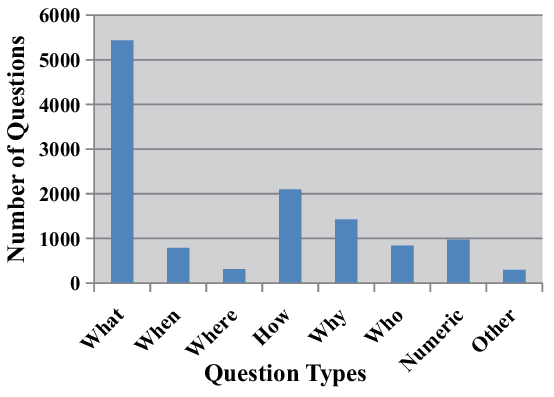}
\caption{The number of questions that belong to each question types in CMedMRC.}
\label{figure_question_type}
\end{figure}

Due to the special characteristics of the Chinese language, the question types cannot be simply classified by prefix words of questions~\citep{DBLP:conf/emnlp/RajpurkarZLL16}.
Here, we manually define 8 common question types in the user annotation interface. The statistics of each question type are shown in Figure~\ref{figure_question_type}. The first seven question types usually correspond to special medical answers. For example, the \textit{What} type refers to a question on the name of a drug or a disease, which accounts for more than half of the dataset. A third of the questions belong to the types of \textit{How} and \textit{Why}. The statistics of answer types are also shown in Table~\ref{answer_type_table}. The proportions of \textit{Noun Phrase} and \textit{Description} types are relatively large.
The results are consistent with Figure~\ref{figure_question_type}, since most of \textit{What} questions need to be answered with the above two answer types.
Table \ref{average_length_data} shows the text length of four input data.


\begin{figure}
\centering
\includegraphics[height=4cm, width=7.5cm]{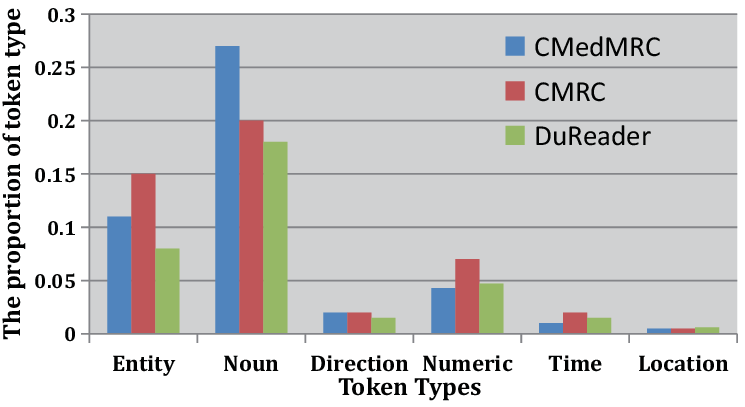}
\caption{Proportions of entities and five frequently appearing POS tags in three Chinese MRC datasets.}
\label{figure_part_of_speech}
\end{figure}

\subsection{Analysis of Domain Knowledge}
We further analyze to what degree there exists domain knowledge in CMedMRC, in terms of medical entities and other terms. In this study, we employ the POS and NER toolkits\footnote{We use jieba toolkit with additional medical term dictionaries. See \url{https://pypi.org/project/jieba/}.} to tag medical entities and terms from 100 samples in the development set of CMedMRC. We also compare the statistics against those of two other Chinese MRC datasets, namely CMRC~\citep{DBLP:conf/emnlp/CuiLCXCMWH19} and DuReader~\citep{DBLP:conf/acl/HeLLLZXLWWSLWW18}.
The proportions of entities and five frequent POS tags in the three datasets are summarized in Figure~\ref{figure_part_of_speech}.
Comparing to the other two open-domain datasets, the proportion of entities in CMedMRC is very high (\textbf{11\%}).
In addition, the proportion of nouns (\textbf{27\%}) is much higher than the other four POS tags in CMedMRC.
The most likely cause is that existing models have difficulty recognizing all the medical terms, and treat them as common nouns.
Among the three Chinese datasets, CMedMRC has the largest proportion (\textbf{38\%}) of nouns and entities.
Therefore, it is difficult for pre-trained language models to understand so many medical terms without additional medical background knowledge.

\section{Experimental Settings}
\label{experimental_result}
\subsection{Medical Knowledge Base and Corpora}
The underlying medical knowledge base is constructed by DXY, containing 44 relation types and over 4M relation triples.
The KGs embedding trained by TransR \citep{DBLP:conf/aaai/LinLSLZ15} on DXY-KG \footnote{\url{https://portal.dxy.cn/}} containing 152,508 entities.
In knowledge retrieval, the threshold of overlapped tokens is set to half of its own length.
The medical pre-training corpora used in ERNIE-THU\citep{DBLP:conf/acl/ZhangHLJSL19} contains 5,937,695 text segments with 3,028,224,412 tokens (4.9 GB) after pre-processing.

\subsection{Additional Training Details}
In average, the training time for DrQA, BERT\_base, MC-BERT, KT-NET, ERNIE, KMQA and CMedBERT takes 10, 16, 16, 27, 29, 28 and 25 minutes per epoch on a TiTAN RTX GPU. All the models are implemented by the PyTorch deep learning framework \footnote{https://pytorch.org/}.

\end{CJK}
\end{document}